\documentclass[10pt,a4paper]{article}

\usepackage{amssymb}
\usepackage{orcidlink}
\usepackage{amsmath}
\usepackage{enumitem}
\usepackage{caption, subcaption}
\usepackage{multirow}
\usepackage{booktabs}
\usepackage{pifont}
\usepackage{xcolor}
\usepackage{graphicx}
\usepackage{authblk}   % for multiple authors with affiliations
\usepackage{dblfloatfix}
\usepackage{cleveref}
\usepackage[acronym, automake, nonumberlist]{glossaries-extra}
\makeglossaries
\setabbreviationstyle[acronym]{long-short}

\usepackage{pgfplots}
\pgfplotsset{compat=1.18}

\newacronym{mlp}{MLP}{Multi-layer Perceptron}
\newacronym{swap}{SWaP}{Size, Weight and Power}
\newacronym{uav}{UAV}{Unmanned Aerial Vehicle}
\newacronym{vot}{VOT}{Visible Object Tracking}
\newacronym{ekf}{EKF}{Extended Kalman Filter}
\newacronym{ros2}{ROS 2}{Robot Operating Software~2}
\newacronym{pmf}{PMF}{Probability Mass Function}
\newacronym{imu}{IMU}{Inertial Measurements Units}
\newacronym{kf}{KF}{Kalman Filter}
\newacronym{ltp}{LTP}{Long-Term evaluation Protocol}
\newacronym{dsp}{DSP}{Down-Sampling evaluation Protocol}

\newacronym{fps}{FPS}{Frame Per Second}
\newacronym{sr}{SR}{Success Rate}
\newacronym{pr}{PR}{Precision Rate}
\newacronym{msr}{MSR}{Maximum Success Rate}

\newacronym{mata}{MATA}{Modular Asynchronous Tracking Architecture}
\newacronym{eop}{EOP}{Embedded-Oriented evaluation Protocol}
\newacronym{nt2f}{NT2F}{Normalize Time to Failure}
\newacronym{ewma}{EWMA}{Exponentially Weighted Moving Average}

\usepackage[
    left=1.6cm,
    right=1.6cm,
    top=1.9cm,
    bottom=2.54cm,
    columnsep=0.5cm
]{geometry}
\usepackage[utf8]{inputenc}   % For pdfLaTeX; skip if you use XeLaTeX/LuaLaTeX
\usepackage[T1]{fontenc}
\usepackage{textcomp}
\usepackage{lmodern}
\usepackage{hyperref}
\hypersetup{
	colorlinks=true, % colore les liens au lieu de les encadrer
	pdfstartview=FitV, % ouvre le PDF de façon à ce qu'il prenne la taille verticale de l'écran
	urlcolor=green, % choix de la couleur des liens URL
	linkcolor=red, % choix de la couleur des liens internes (table des matières, etc.)
	citecolor=blue
}
\newcommand{\cf}{\emph{cf.}\ }
\newcommand{\cmark}{\textcolor{green}{\ding{51}}}%
\newcommand{\xmark}{\textcolor{red}{\ding{55}}}%

\title{Architecture and evaluation protocol for transformer-based visual object tracking in UAV applications} 
\author[1,2,3]{Augustin Borne\thanks{Corresponding author: augustin.borne@uha.fr}~\orcidlink{0009-0000-0444-2074}}
\author[1]{Pierre Notin~\orcidlink{0009-0001-9746-3823}}
\author[1]{Dr. Christophe Hennequin~\orcidlink{0009-0004-4776-4013}}
\author[1]{Dr. Sebastien Changey~\orcidlink{0000-0002-6490-4653}}
\author[2]{Dr. Stephane Bazeille~\orcidlink{0000-0001-9670-9863}}
\author[2]{Prof. Christophe Cudel~\orcidlink{0000-0003-0356-2300}}
\author[3]{Prof. Franz Quint~\orcidlink{0000-0002-8757-6883}}

\affil[1]{French-German Research Institute of Saint-Louis, Saint-Louis, France}
\affil[2]{Université de Haute-Alsace, Mulhouse, France}
\affil[3]{Karlsruhe University of Applied Sciences, Karlsruhe, Germany}
\date{}

\begin{document}

\maketitle

% Author: Please give full first and last names for authors and include * after the name of all corresponding authors
% Abstract should be written in the present tense and impersonal style (i.e., avoid we), and be at most 200 words long
\begin{abstract}
\noindent Object tracking from Unmanned Aerial Vehicles (UAVs) is challenged by platform dynamics, camera motion, and limited onboard resources. Existing visual trackers either lack robustness in complex scenarios or are too computationally demanding for real-time embedded use. We propose an Modular Asynchronous Tracking Architecture (MATA) that combines a transformer-based tracker with an Extended Kalman Filter, integrating ego-motion compensation from sparse optical flow and an object trajectory model. We further introduce a hardware-independent, embedded oriented evaluation protocol and a new metric called Normalized Time to Failure (NT2F) to quantify how long a tracker can sustain a tracking sequence without external help. Experiments on UAV benchmarks, including an augmented UAV123 dataset with synthetic occlusions, show consistent improvements in success rate and NT2F metrics. A ROS 2 implementation on an NVIDIA Jetson AGX Orin confirms that the proposed evaluation protocol better represents embedded performances. Across the tested methods, the discrepancy with the embedded baseline is on average $6.4\times$ smaller for success rate and $4.7\times$ smaller for NT2F compared to the LTP protocol.
\end{abstract}
\vspace{5pt}


\section{Introduction}
\label{sec:intro}

For lot of \gls{uav} applications, visual tracking of objects is an essential task. However, this is difficult due to high platform dynamics conducting to significant camera motion, frequent object occlusions coming from obstacles, and scale/perspective changes from highly varying distance to the object of interest. In addition, the entire perception pipeline must run in real time on \gls{swap} constrained embedded hardware.

\gls{vot} refers to following an identified object over a sequence of video frames. The search is usually restricted to a local region for computational efficiency. However, this approach struggles with fast object motion, occlusions, and camera movements. Prior to the emergence of deep-learning-based trackers, correlation-filter methods such as MOSSE~\cite{5539960} and KCF~\cite{10.1109/TPAMI.2014.2345390} were widely used due to their computational efficiency. While computationally efficient, they are sensitive to occlusions and large deformations. The advent of deep learning introduced Siamese network-based trackers, e.g SiamFC~\cite{10.1007/978-3-319-48881-3_56} and  SiamRPN~\cite{8579033}. They significantly improved accuracy by learning robust appearance representations, yet they can struggle with long-term occlusions and significant scale changes. More recently, transformer-based architectures have been adapted for \gls{vot}. They model global context and improve resilience to difficult challenges such as occlusions, deformations, and background clutter. However, they are computationally heavy. Notable models include TransT~\cite{chen2021transformertracking}, OSTrack~\cite{ye_joint_2022}, and SeqTrack~\cite{10203645}. Recent approaches also explore prompt-based and foundation model trackers, leveraging textual or visual prompts to enable flexible tracking, including PIVOT~\cite{10855520} or ChatTracker~\cite{10.5555/3737916.3739157}, however prompt-based models are often too large for embedded real-time deployment. 

Despite progress in \gls{vot}, key challenges remain. Standard trackers often ignore the temporal dimension, and struggle with rapid object motion, occlusions, or camera-induced movement. Accuracy and efficiency trade-offs are significant: transformers are precise but slow, correlation filters are fast but less accurate, and Siamese networks lie in between. Moreover, current evaluation protocols assume a single tracker operating at a fixed processing rate and do not account for systems composed of multiple processing blocks running at different frequencies. As a result, they fail to properly evaluate architectures that leverage asynchronous components to achieve the best possible performance under real-time constraints. These limitations highlight that there is still substantial room for improvement in designing and evaluating tracking solutions that are both accurate and able to run in real-time on embedded hardware on an \gls{uav}.

In this study, we address these challenges by proposing a \gls{mata}, which combines a transformer-based tracker, an explainable score mechanism and an \gls{ekf}~\cite{10.5555/897831}. This design leverages the high accuracy of vision transformers while benefiting from the \gls{ekf}’s real-time estimation capabilities on embedded devices. The \gls{ekf} integrates camera ego-motion compensation, estimated via a lightweight sparse Lucas-Kanade optical flow~\cite{10.5555/1623264.1623280}, with an object trajectory model.  We validate the the proposed architecture method on UAV datasets (such as UAV123~\cite{10.1007/978-3-319-46448-0_27} and VTUAV~\cite{9879218}). In many \gls{uav} applications, the object may be subject to partial or complete occlusions. Thus, evaluating a tracking system’s ability to produce coherent prediction in such situation is important, as successfully passing an occlusion is a strong indicator that the tracker considers system dynamics. However, most existing datasets contain only a limited number of occlusion events, which makes this aspect difficult to evaluate in a controlled manner. To address this limitation, we developed an augmentation tool that synthetically generates smooth occlusions of varying shapes and sizes following the object trajectory. This allows trackers to be evaluated more thoroughly under occlusion scenarios. Using this tool, we created an augmented version of the UAV123 dataset, named UAV123+occ. We use this augmented dataset to further evaluate the proposed tracking architecture.

To evaluate multi-module tracking systems and obtain performance metrics representative of real deployment scenarios on embedded systems, we introduce the \gls{eop}, a hardware-independent evaluation protocol that simulates onboard processing delays. It models asynchronous data flows between modules, allowing each component to operate at its own frequency. This modular protocol enables fair cross-platform comparison while assessing both temporal robustness and real-time feasibility. The evaluation protocol representativity is validated by comparing its results to a \gls{ros2} implementation on a Nvidia Jetson AGX Orin~\cite{nvidia_orin_launch_2022}. 

In many real-world applications, trackers may loose object due to harsh conditions, and re-initializing a tracker after a failure may not be possible. In these conditions, trackers must focus on temporal reliability capabilities. Thus, it is important to quantify how long a tracker can reliably maintain a target without external intervention. To address this, we formalize the \gls{nt2f} metric, which measures the duration for which a tracker can successfully follow an object before a tracking failure occurs. This metric is used to assess the contribution of the our framework.
 
\noindent The key contributions of this work are:

\begin{enumerate}
    \item A Modular Asynchronous Tracking Architecture (\gls{mata}) that enhances robustness in difficult conditions, including occlusions.
    \item An embedded oriented protocol (\gls{eop}) to perform hardware-independent evaluation on tracking method.
    \item The design of a temporal reliability evaluation metric (\gls{nt2f})
    %\item An augmentation tool to generate synthetic occlusions on \gls{vot} datasets
\end{enumerate}

\section{MATA: a Modular Asynchronous Tracking Architecture}
\label{sec:methodology}

\begin{figure*}[!ht]
  \centering
  \includegraphics[width=0.9\linewidth]{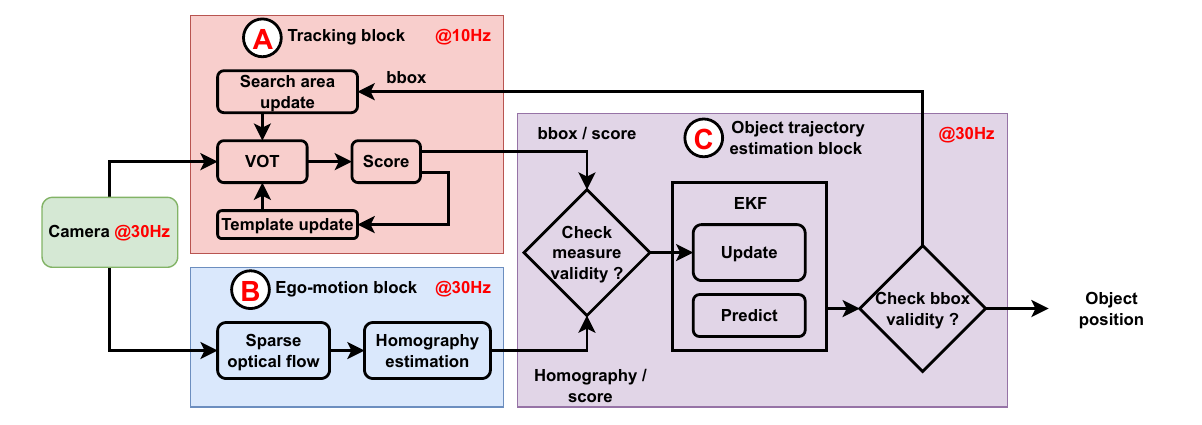}
  \caption{Overview of the Modular Asynchronous Tracking Architecture: Example with tracker running at 10~Hz, estimation filter at 30~Hz and ego motion at 30~Hz.}
  \label{fig:mta}
\end{figure*}

In an image sequence, an object’s apparent motion comes from two sources: the object’s own movement and the motion of the camera. In order to cope with this situation, we propose a tracking framework called \gls{mata}, which organizes independent, interchangeable blocks around a transformer-based tracking algorithm. This framework models these two effects separately and combines them in an estimation-filter block to produce bounding boxes at real-time rates. As shown in \cref{fig:mta}, the architecture includes three main modules: 

\begin{enumerate}[label=\Alph*.]
\item a visual tracking block based on a vision-transformer model that measure object position in an image
\item a ego-motion compensation block that estimates the camera’s ego-motion
\item an estimation block that predict an object position from a dynamic model, exploiting position predicted by \gls{vot} and camera motion measurement
\end{enumerate}

\noindent Thanks to its modular design, any block in \gls{mata} can be replaced or upgraded with block of the same functionality. Furthermore, the architecture could be easily extended with additional components as needed.

\subsection{Explainable VOT}
\label{sub_sec:score}

\begin{figure}[!b]
\centering
\begin{tikzpicture}
\begin{axis}[
    width=0.8\linewidth,
    height=6cm,
    xlabel={Index in feature map},
    ylabel={Probability},
    grid=major,
    legend style={anchor=north east}
]

% PMF curve
\addplot[
    thick,
] coordinates {
(0, 1.6312707884935662e-05)
(1, 2.917691381298937e-05)
(2, 3.663406823761761e-05)
(3, 2.617945574456826e-05)
(4, 1.5215089661069214e-05)
(5, 8.556235115975142e-06)
(6, 1.0129427209903952e-05)
(7, 1.4137965990812518e-05)
(8, 1.2007983968942426e-05)
(9, 8.929625437303912e-06)
(10, 5.633016826323001e-06)
(11, 1.1348717634973582e-05)
(12, 3.4692668123170733e-05)
(13, 3.6835859646089375e-05)
(14, 3.0779523513047025e-05)
(15, 2.1886147806071676e-05)
(16, 1.7552956705912948e-05)
(17, 2.3174705347628333e-05)
(18, 2.3109152607503347e-05)
(19, 1.605867873877287e-05)
(20, 1.4885637938277796e-05)
(21, 8.571435500925872e-06)
(22, 7.189117241068743e-06)
(23, 1.1377214832464233e-05)
(24, 7.404052212223178e-06)
(25, 3.887303137162235e-06)
(26, 2.068311005132273e-06)
(27, 1.6307770920320763e-06)
(28, 3.581290911824908e-06)
(29, 5.514773420145502e-06)
(30, 8.84907512954669e-06)
(31, 1.8577500668470748e-05)
(32, 0.00015707449347246438)
(33, 0.005195517558604479)
(34, 0.0677393302321434)
(35, 0.2822824716567993)
(36, 0.1629745066165924)
(37, 0.06731352210044861)
(38, 0.0934734046459198)
(39, 0.22315223515033722)
(40, 0.07792890816926956)
(41, 0.01502701360732317)
(42, 0.002148026367649436)
(43, 0.0006569622782990336)
(44, 0.00025484946672804654)
(45, 0.00014544730947818607)
(46, 5.459509338834323e-05)
(47, 3.314510468044318e-05)
(48, 1.671043355599977e-05)
(49, 3.9058435504557565e-05)
(50, 4.1428393160458654e-05)
(51, 3.304464553366415e-05)
(52, 2.2469806935987435e-05)
(53, 1.3768252756563015e-05)
(54, 1.308589617110556e-05)
(55, 2.3102616978576407e-05)
(56, 2.5139663193840533e-05)
(57, 2.523697003198322e-05)
(58, 1.803646591724828e-05)
(59, 2.3896282073110342e-05)
(60, 4.445617014425807e-05)
(61, 5.111200880492106e-05)
(62, 5.156655970495194e-05)
(63, 4.098819772480056e-05)
(64, 1.689799319137819e-05)
(65, 2.5663417545729317e-05)
(66, 1.7944314095075242e-05)
(67, 1.0613265658321325e-05)
(68, 7.2012358032225166e-06)
(69, 6.4495579863432795e-06)
(70, 8.674101991346106e-06)
(71, 2.920157385233324e-05)
(72, 3.16307159664575e-05)
(73, 2.1220317648840137e-05)
(74, 9.62263857218204e-06)
(75, 7.401869879686274e-06)
(76, 1.1912809895875398e-05)
(77, 1.3303025298228022e-05)
(78, 1.496724689786788e-05)
(79, 9.348823368782178e-06)
(80, 6.541028142237337e-06)
(81, 1.1876804819621611e-05)
(82, 9.146912816504482e-06)
(83, 1.084479117707815e-05)
(84, 1.139092000812525e-05)
(85, 1.0858593668672256e-05)
(86, 1.370686550217215e-05)
(87, 2.521819078538101e-05)
(88, 1.610670005902648e-05)
(89, 1.6354491890524514e-05)
(90, 1.2149018402851652e-05)
(91, 1.2330217941780575e-05)
(92, 2.282576315337792e-05)
(93, 2.4934259272413328e-05)
(94, 1.5388532119686715e-05)
(95, 2.631839015521109e-05)
};
\addlegendentry{PMF}

% AUC shaded region
\addplot[
    fill=blue!30,
    draw=none,
    area legend
] coordinates {
(32, 0.00015707449347246438)
(33, 0.005195517558604479)
(34, 0.0677393302321434)
(35, 0.2822824716567993)
(36, 0.1629745066165924)
(37, 0.06731352210044861)
(37,0)
(32,0)
};
\addlegendentry{AUC}

% Peak vertical line
\addplot[
    red,
    dashed,
    thick
] coordinates {
(35,0)
(35, 0.2822824716567993)
};
\addlegendentry{Peak}

\end{axis}
\end{tikzpicture}
\caption{Probability mass function with detected peak and AUC region. }
\label{fig:aux_score}
\end{figure}
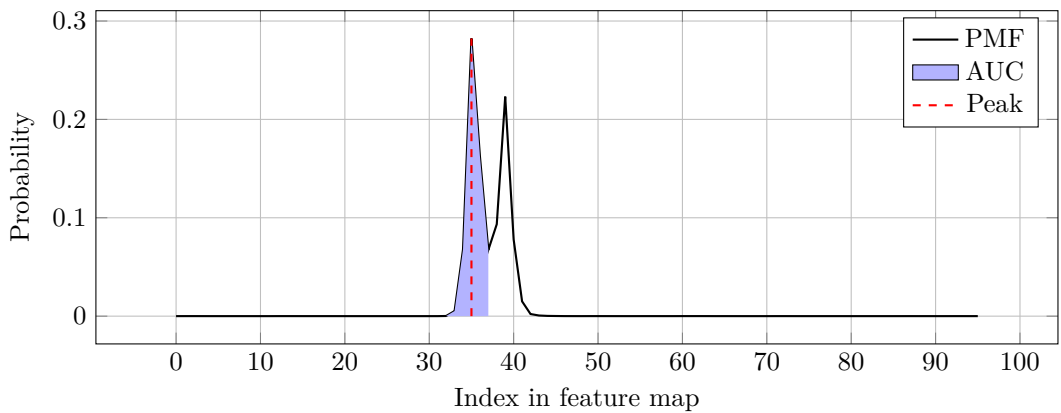

The first block of our the architecture is the tracking block (Block~A in \cref{fig:mta}), which is centered around a Vision Transformer-based \gls{vot} model. The objective of this block is to exploit the visual information provided by the image sensor to estimate the position of the object and to feed this measurement to Block~C, the object trajectory estimation block. A key requirement is to determine whether the estimated object position is reliable, in order to avoid providing erroneous measurements to the estimator when the tracker has lost the object. For this purpose, a confidence score derived from the tracker output is required. Ideally, this score should represent a probability within the interval $[0,1]$, which improves interpretability and facilitates the decision process regarding the validity of the measurement.

Two Vision Transformer-based trackers are considered in this work: MixFormerV2~\cite{10.5555/3666122.3668683} and OSTrack~\cite{ye_joint_2022}. Both models produce confidence scores within $[0,1]$ respectively using softmax and sigmoid activation functions. However, scores produced through sigmoid-based outputs may suffer from reduced interpretability due to their non-normalized nature, whereas softmax-based outputs can be interpreted more directly as probabilities. In the case of MixFormerV2, the confidence score is produced by a dedicated \gls{mlp} head, which makes its interpretation less transparent and decouples it from the features directly used to estimate the bounding box. To address these limitations, we propose a new confidence score derived directly from the softmax output used for bounding box prediction. This formulation can be naturally integrated with the MixFormerV2 output that can be interpreted as a probability mass function (PMF). Such formulation cannot be applied to OSTrack due to architectural differences. Consequently, when using OSTrack within the \gls{mata} framework, the original tracker confidence output is used without modification.

The bounding box produced by the MixFormerV2 model is defined by its top-left and bottom right corners coordinates noted respectively ($x_{tl}$, $y_{tl}$) and ($x_{br}$, $y_{br}$). Let $X_{x_{tl}}$, $X_{y_{tl}}$, $X_{x_{br}}$ and $X_{y_{br}}$ be discrete random variables. Their \gls{pmf} are defined as $p_i(x) = \mathbb{P}(X_i = x)$ with $i \in \{ x_{tl}, y_{tl}, x_{br}, y_{br}\}$ and $x \in [0, N]$ where $N$ is the number of discrete bins in the \gls{pmf}. In MixFormerV2, each side of the bounding box is represented as a discrete probability distribution over $N$ positions rather than predicting pixel coordinates directly. This allows the network to model uncertainty in the box location and improves robustness under challenging conditions, such as occlusions or fast motion. An example of a \gls{pmf} output from MixformerV2 model is shown in \cref{fig:aux_score}.

Based on this principle, we propose a confidence score that leverages the \gls{pmf} produced by the MixFormerV2 model. For each \gls{pmf}, the peak position is defined as $k_i^* = \arg\max_{k} p_i(k)$ and for each coordinate $i$, we compute a score $S_i$ as defined in \cref{eq:score}.

\begin{equation}
\label{eq:score}
\begin{aligned}
S_i &= \sum_{k = k_{\min,i}}^{k_{\max,i}} p_i(k) \\
k_{\min,i} &= \min\!\left(k_i^* - \tfrac{\alpha N}{2},\, 1\right)\\
k_{\max,i} &= \max\!\left(k_i^* + \tfrac{\alpha N}{2},\, N\right)
\end{aligned}
\end{equation}

\noindent  A window of width $\alpha N$, where $\alpha$ is a hyperparameter, centered at $k_i^*$, is defined. The parameter $\alpha$ was empirically determined to be $0.03$ via a grid search procedure. The score for each coordinate is computed as the sum of probabilities within this window (The AUC area in \cref{fig:aux_score}). The bounds $k_{\min,i}$ and $k_{\max,i}$ define the window limits and handle boundary effects at the edges of the feature map. The final confidence score is defined by the lowest confidence score over the four coordinate scores, $S = \min_{i} S_i$. By utilizing these \gls{pmf}, this scoring method quantifies how concentrated the predicted probabilities are around their modes, providing an interpretable and spatially grounded confidence estimate for the model output. While the score is applied in 1D along each bounding box coordinate due to the special 1D output format of MixFormerV2, it can be easily extended to 2D heatmaps by replacing the linear window around the mode with a square window over the spatial grid, provided the heatmap resolution is sufficient.

Our scoring method is more interpretable than MixFormerV2’s original \gls{mlp}-based score by relying on the predicted probability distributions. Such explainability is critical for UAV applications that require reliable confidence estimates for autonomous decision-making. However this new confidence measure requires a model output with sufficient resolution. Thus, it could not be used for any models and in the \gls{mata} system, the original confidence score is replaced only when it's possible. 

Depending on the underlying \gls{vot} algorithm, the tracker may rely on fixed, online, or hybrid template strategies. In this work, no modification is applied to trackers that use a fixed-template mechanism. However, for trackers employing online or hybrid template updates, the update mechanism is modified to follow a threshold-based strategy conditioned on the proposed confidence score. Specifically, the template is updated only when the confidence score exceeds a predefined threshold $\tau_\text{template}$. This prevents the tracker from incorporating corrupted object appearances when the tracking confidence is low, thereby improving robustness. In practice, this modification is applied to MixformerV2, which uses a hybrid template strategy, while OSTrack remains unchanged.

\subsection{Camera movement compensation}
\label{subsec:cam_motion}

The second block of our architecture is the ego-motion compensation block (Block~B in \cref{fig:mta}), which mitigates the motion induced by the camera, allowing the true dynamics of the object of interest to be isolated. It takes as input the image at time step $k$ and estimates the transformation between images $k-1$ and $k$, accompanied by a confidence measure for this estimate, ideally expressed in the form of a covariance matrix.

Ego‑motion can be estimated by fusing visual and \gls{imu} data, by using onboard GNSS measurements, or by relying solely on image sequences. However, inertial measurements accumulate integration drift over time and therefore usually require sensor fusion to maintain reliable long-term ego-motion estimation. Similarly, GPS measurements are typically available at relatively low update rates (e.g., around 10 Hz), which may be insufficient for accurately capturing the fast dynamics of high-speed UAVs. Furthermore, GPS signals may be unavailable or significantly degraded in certain operational environments, such as urban canyons, dense forests, or indoor settings. Consequently, and to focus the scope of this work on visual tracking, in a first approach we restrict our methods to image‑based approaches only. 

Camera motion is approximated with geometric models of varying complexity~\cite{Hartley_Zisserman_2004}, ranging from translation and similarity transforms to affine and full projective homographies. A practical and lightweight solution, used in OpenCV~\cite{opencv_library}, tracks sparse features detected with standard corner or keypoint detectors~\cite{Shi_Tomasi_1994, Rublee_2011} using pyramidal Lucas–Kanade optical flow~\cite{Bouguet1999PyramidalIO}, and estimates an affine transform between frames. Each frame is then warped accordingly, with slight cropping or zooming to hide border artifacts.

In our application setting, onboard computation is severely constrained, making feature-based estimation impractical because feature detection and matching dominate runtime of the ego-motion compensation pipeline. To estimate inter-frame camera motion, we use a two-stage approach combining sparse optical flow and transformation optimization. 

To reduce computational cost of the sparse optical flow part, we sample points on a regular grid rather than performing feature detection. We then use the classical approach with the pyramidal Lucas–Kanade algorithm to track points, producing correspondences $\{(\mathbf{p}_i, \mathbf{p}'_i)\}_{i=1}^N$, where $\mathbf{p}_i$ and $\mathbf{p}'_i$ respectively denote the positions of the $i$-th point in the current and next frame. Grid points located in low-contrast regions, where optical flow cannot be reliably estimated, are discarded.

To better capture \gls{uav} camera motion, which can include significant perspective changes at low altitude, we model the 2D inter-frame relationship using an homography $\mathbf{H} \in \mathbb{R}^{3\times3}$, with $g$ the homography projection as detailed in \cref{eq:homography}.

\begin{equation}
\label{eq:homography}
\begin{aligned}
g(x, y) &= \begin{bmatrix}x'\\y'\end{bmatrix} \text{ with }
s\begin{bmatrix}x'\\y'\\1\end{bmatrix} \sim H \begin{bmatrix}x\\y\\1\end{bmatrix}\\
H &=\begin{bmatrix}
h_{11} & h_{12} & h_{13} \\
h_{21} & h_{22} & h_{23} \\
h_{31} & h_{32} & h_{33}
\end{bmatrix}
\end{aligned}
\end{equation}

\noindent We use RANSAC~\cite{10.1145/358669.358692} to robustly estimate the transformation by identifying the subset of points whose optical flow is consistent with a common motion model. The transformation is then computed using only the inlier correspondences. We compute the covariance of the estimated homography using the first-order approximation of non-linear parameter uncertainty, as described in Hartley and Zisserman~\cite{hartley_multiple_2018}, Result 5.10. This covariance matrix quantifies the uncertainty in the estimated transformation and is used as a confidence score for the estimated inter-frame ego-motion in downstream tasks.

\subsection{Object trajectory estimation}
\label{subsec:target_traj_est}

The third and final block of the architecture is the object trajectory estimation block (Block C in \cref{fig:mta}).
It aims to estimate object position using a motion model alongside sensor measurements to iteratively compute the best guess of the object position over time. 

More precisely, its objective is to provide an estimation of the object position in the current video frame, using a state transition function that describe the assumed object dynamic behavior. The predicted model states are updated by the available measured data at current frame such as the camera inter-frame ego-motion (from block B) and object position (from block A). This approach brings greater temporal reliability to the architecture by establishing asynchronous capabilities, with the estimator computing object estimated position at a faster rate than the measurement from the tracker. Furthermore, the estimation filter can yields coherent estimations when the tracker cannot provide a reliable measure of object position either while it is unavailable due to computing time, or because the object becomes unobservable (e.g., due to occlusion), requiring the position to be predicted from camera motion estimation and object dynamic model.

Although the estimation problem is non-trivial due to the observed 2D object motion resulting from the combination of two distinct dynamic components: the ego-motion of the camera and the intrinsic motion of the object. Estimation filters are typically used to handle this object tracking problem.
Indeed, in \gls{vot} field, \gls{kf} based~\cite{kalman_filter} approaches have been employed to enhance base trackers, providing higher robustness to occlusion such as mean-shift~\cite{Karavasilis, ISWANTO2017587} and STC~\cite{electronics10010043}, while in multi-object tracking, variants like SORT~\cite{Bewley_2016} improve tracking by combining a \gls{kf} with efficient data association. The BOT-SORT variant~\cite{aharon2022botsortrobustassociationsmultipedestrian} incorporates ego-motion compensation to enhance tracking performance, while OC-SORT~\cite{cao2023observationcentricsortrethinkingsort} introduces an Observation-Centric Re-Update mechanism that uses virtual trajectories to stabilize updates after extended periods of prediction-only tracking.  

In the considered context it is desirable to keep temporal reliability even in complex cases such as occlusions or high dynamics. A typical assumption about the object dynamics is to considered it as a linear model (Constant-Velocity) of the form $p_{k}=p_{k-1}+v_{k-1}$ with $p$ the object position, $v$ the object velocity, and the subscript $k$ indicating the current time step. This approach is particularly suitable when considering short period of time between two time step $k-1$ and $k$, and when the image frame in which is expressed the object position has relatively slow changes with reference to the object dynamics, i.e. when the camera ego-motion dynamic is negligible with reference to the observed object dynamics. However, in \gls{uav} applications this assumption can not be maintained. Thus, to alleviate this problematic we propose to measure the frame transformation between two image frames through an homography matrix (block B, \cref{subsec:cam_motion}) and use it to project the object constant velocity model from the frame $k-1$ to the frame $k$. This more precise description, lead to a non-linear model which isn't suitable for \gls{kf} estimation. Thus, we employ an \gls{ekf}~\cite{bryson_applied_1975} that offer the benefit of handling the inherent nonlinearity of the observed system, by linearizing the model (state transition function) around the current estimate, while keeping the simplicity of \gls{kf} and a satisfying computing time.
This serves as a proof of concept demonstrating that combining an estimation filter with a vision-transformer-based \gls{vot} model can improve tracking performance. 

The filter state and measurements vectors are defined in \cref{eq:ekf_x_z}. The state vector is combining the bounding box coordinates $\mathbf{X}_{tl} = \begin{bmatrix}x_\text{tl}, y_\text{tl}\end{bmatrix}$, $\mathbf{X}_{br} = \begin{bmatrix}x_\text{br}, y_\text{br}\end{bmatrix}$, the homography parameters $\mathbf{X}_{H} = \begin{bmatrix}h_\text{11}, h_\text{12}, ..., h_\text{32}, h_\text{33}\end{bmatrix}$, and the bounding box center velocity $\mathbf{X}_{v} = \begin{bmatrix}\dot{x}_\text{c}, \dot{y}_\text{c}\end{bmatrix}$. The measurement vector follows the same partitioning and is composed only of the observable quantities $\mathbf{Z}_{tl} = \begin{bmatrix}x_\text{tl}, y_\text{tl}\end{bmatrix}$, $\mathbf{Z}_{br} = \begin{bmatrix}x_\text{br}, y_\text{br}\end{bmatrix}$ and $\mathbf{Z}_{H} = \begin{bmatrix}h_\text{11}, h_\text{12}, ..., h_\text{32}, h_\text{33}\end{bmatrix}$. Since no direct measurements of the bounding box center velocity are available, $\mathbf{X}_{v}$ are estimated through the prediction step of the filter.

\begin{equation}
\label{eq:ekf_x_z}
\begin{aligned}
\mathbf{X} &=
\begin{bmatrix}
\mathbf{X}_{tl} \ \mathbf{X}_{br} \ \mathbf{X}_{H} \ \mathbf{X}_{v}
\end{bmatrix}^\top \\
\mathbf{Z} &=
\begin{bmatrix}
\mathbf{Z}_{tl} \
\mathbf{Z}_{br} \
\mathbf{Z}_{\text{H}}
\end{bmatrix}^\top
\end{aligned}
\end{equation}

\noindent During prediction step, the predicted state vector is computed from the state transition function $f$ that can be split in four parts similarly to state vector $\mathbf{X}$, giving the \cref{eq:ekf_velocity_1}. 

\gls{uav} can have erratic movements. Thus, we are not taking assumptions to model its dynamic and consequently the homography parameters are assumed to be constant between two consecutive time steps $\mathbf{X}_{H,k}=\mathbf{X}_{H,k-1}$. This leaves the projection dependent on the interframe camera ego-motion measurements.
Let $g$ denote the homography projection defined in \cref{subsec:cam_motion}. Let $\mathbf{J_g}(x,y)$ denote the Jacobian of the projection function $g$ with respect to the pixel coordinates $(x,y)$. Both of these functions are dependent on $\mathbf{X}_{H}$ states as described in \cref{eq:homography}. 
Then, the projection of the constant velocity object model from frame $k-1$ to $k$, is realized by coupling the bounding box corners with the camera-induced motion described by the homography projection $g(\cdot)$ and the object velocity state projected in the current image frame through the Jacobian $\mathbf{J_g}$. Finally, when measurement are available, predicted states are refined through the updating step.

\begin{equation}
\label{eq:ekf_velocity_1}
\begin{aligned}
\mathbf{X}_k &= f(\mathbf{X}_{k-1}) = 
\begin{bmatrix}
\mathbf{X}_{tl, k}\\
\mathbf{X}_{br, k}\\
\mathbf{X}_{H, k}\\
\mathbf{X}_{v, k}\\
\end{bmatrix}\\ 
\begin{bmatrix}
\mathbf{X}_{tl, k}\\
\mathbf{X}_{br, k}\\
\mathbf{X}_{H, k}\\
\mathbf{X}_{v, k}\\
\end{bmatrix} &= 
\begin{bmatrix}
\mathbf{g}(\mathbf{X}_{tl,k-1}) + \mathbf{J_g}(\mathbf{X}_{v,k-1}) dt\\
\mathbf{g}(\mathbf{X}_{br,k-1}) + \mathbf{J_g}(\mathbf{X}_{v,k-1}) dt\\
\mathbf{X}_{H,k-1}\\
\mathbf{J_g}(\mathbf{X}_{v,k-1})
\end{bmatrix}
\end{aligned}
\end{equation}

\noindent The \gls{ekf} initial state $\mathbf{X}_0$ is set as the initialization bounding box for $\mathbf{X}_{bb_0}=[\mathbf{X}_{tl_0} \ \mathbf{X}_{br_0}]$, with homography $\mathbf{X}_{H_0}$ set to the identity and bounding box velocity $\mathbf{X}_{v_0}$ set to zero. The process noise covariance $\mathbf{Q}$ and the initial state covariance $\mathbf{P}_0$ are treated as hyper-parameters and have been tuned experimentally, with different values assigned to the bounding box, ego-motion, and velocity components to balance responsiveness and stability.  

Regarding the measurements, they are subject to considerable variation in how accurately they reflect the true quantities, especially the tracking block. In the context of \gls{vot}, a "false positive" is an event that occurs when the tracking model is reporting the position of the tracked object at a location where the object is not actually present. Usually, it happens when the object is occluded or when the model is switching to a distractor, i.e. an object of the same nature as the object of interest. These degenerate measures could generate important drift in the estimated object position if passed to the \gls{ekf}. Consequently, it is desirable to introduce a rejection mechanism prior to the \gls{ekf} that prevents any unsatisfying measurement to perturb the estimation pipeline. This mechanism then allows the measurement noise covariance matrix $\mathbf{R}$ to be set with fixed low values, reflecting the assumed reliability of the retained measurements.

A simple yet effective solution is to define threshold methods, fixing conditions for which a measure is automatically rejected. For each processed frame, the tracker and ego-motion components provide a measurement along with a confidence score. Since both components rely on image data, their outputs are highly dependent on scene conditions and can be significantly variable. The confidence score reflects, to some extent, the reliability of each measurement, making them ideal candidates for being values used in the threshold conditions. Thus, to update the filter state using the tracking block (block A) output, we maintain an \gls{ewma} noted $S_k$ of the predicted confidence score at frame $k$ as detailed in \cref{eq:ewma_score} where $P_k$ is the confidence score of the \gls{vot} model at frame $k$ and $\alpha_{ewma}$ is the \gls{ewma} smoothing factor. If $S_k$ falls below a threshold noted $\tau_\text{ewma}$ or the current score drops significantly relative to $S_k$ (by more than $\tau_\text{diff}$), the tracker measurement is discarded; otherwise, it is used to update the target state. This threshold is preferred to a simple threshold to avoid dropping measurement frame to frame and rather rejecting measurement when the mean confidence in consecutive frame is quite low. Still, we keep a dynamic rejecting system if the confidence score drop significantly from one frame to another, indicating a probable object loss

\begin{equation}
    \label{eq:ewma_score}
    S_k = \alpha_{ewma}P_k + (1 - \alpha_{ewma})S_{k-1}
\end{equation}

\noindent With the same idea, the inter-frame transformations from the ego-motion block (block B) are used only if their covariance is below a maximum threshold $\tau_\sigma$. As each block (A and B) operate at different frequencies, the filter performs partial updates whenever new measurements arrive. 

However, despite measurement confidence considerations, this strategy remains sensitive to false positives associated with high measure confidence. Making the false measured quantity transparent with regard to the threshold conditions, consequently degrading the \gls{ekf} estimation quality. This is a common issue for any \gls{vot} model and is outside the scope of this work, as we do not modify the model itself in this study.

Finally, the output of the object trajectory estimation block (block C) represents the output of \gls{mata}, giving the estimated object position in the current frame $k$. %\textcolor{red}{When the tracker giving estimated position for frame $k-T$ ?} 
This predicting capability allows the estimated object position to be used as feedback to the tracking block (block A) to define the center of the tracker search region. By directing the search area according to the estimated object trajectory, this mechanism ensures that, even if the tracker runs at a lower frame rate, the combined motion of the camera and the object of interest is accounted for when providing the next image to the model. Noticeably helping the tracking algorithm to retrieve the object after an occlusion.

\section{Experiment}
\label{sec:experiment}

\subsection{Embedded oriented evaluation protocol}
\label{subsec:new_eval}

In \gls{vot}, an evaluation protocol defines how a method is run on a dataset. In the literature, several widely used protocols exist. First, in the short-term protocol, the tracker is reinitialized after each failure, focusing on accuracy in nominal conditions. Second, the \gls{ltp}, dominant in the literature, initializes the tracker once and evaluates continuous localization and target-presence confidence. Third, to account for real-time constraints, VOT2021~\cite{9607833} introduced a real-time variant, in which frames are processed at a fixed rate (e.g., 20~Hz), and missing predictions are completed using a zero-order hold. Finally, one option is to evaluate trackers by applying down-sampling to the input frames, yielding what we refer to as the \gls{dsp}. In this protocol, frames are uniformly sub-sampled (e.g., keeping every $n$th frame), resulting in sequences with lower effective frame rates. Such sub-sampling has been used in evaluation studies to investigate tracker performance at different temporal resolutions by discarding intermediate frames~\cite{galoogahi_need_2017}.

All these protocols assume a monolithic tracker that directly processes image frames, which is unrealistic for embedded or robotic systems. In practice, a full processing pipeline consists of multiple interconnected modules, each operating at its own frequency. Real-time behavior therefore depends not only on algorithmic complexity but also on the hardware platform, meaning that the same tracker can behave very differently on a GPU-powered system versus a lightweight embedded device. This implies that an evaluation protocol should be independent of the hardware to ensure reproducibility, while still taking processing frequency into account. This is not the case for the real-time protocol, which reflects real-time tracking performance as a function of the underlying hardware. Consequently, when sufficient computational resources are available such that all frames are processed in real time (superior to image frequency), this protocol becomes equivalent to the \gls{ltp}. The same limitation applies to the \gls{dsp} protocol. While \gls{dsp} is easy to implement and represents a step toward assessing frame rate effects, it remains insufficient for real-time evaluation, as simply down-sampling the input does not account for the actual processing time of the tracker.

To bridge the gap between development, benchmarking, and deployment, we propose an Embedded Oriented Protocol (\gls{eop}) that mimics an onboard processing behavior. It evaluates tracking using independent blocks with different update rates and optional one-period latency to simulate block processing time (e.g., a 5 Hz tracker on a 30 Hz stream processes every sixth frame and outputs one step later). Blocks communicate via a shared structure, with missed updates being either propagated or skipped according to each block rules. The frequency and output delay of each block are fully tunable by hand and act as hyper-parameters, independent of hardware or software. This allows replicating the same processing conditions across different platforms and implementations. By design, it is modular, enabling new components to be added easily. 

\begin{figure*}[!t]
  \centering
  \begin{subfigure}[c]{\textwidth}
    \centering
    \includegraphics[width=0.9\textwidth]{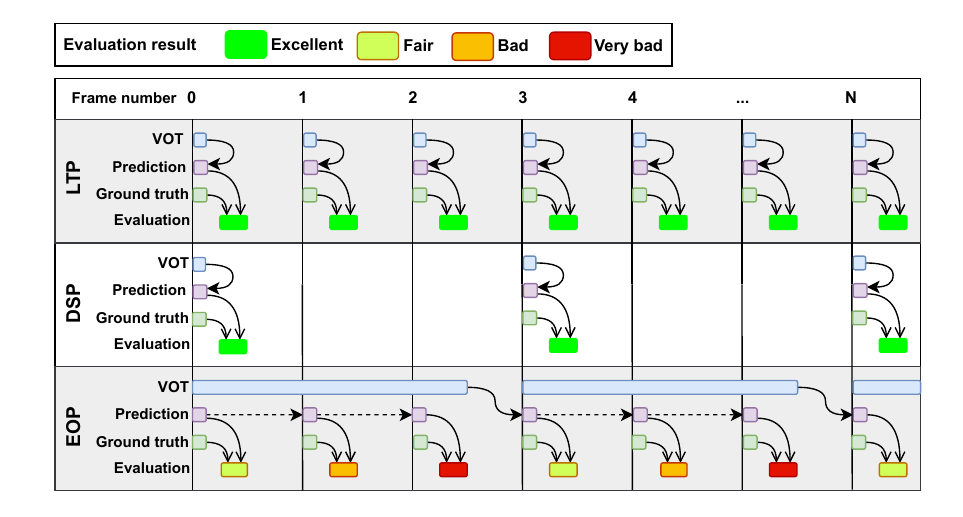}
    \caption{Comparison of evaluation protocols: the long-term protocol (LTP) commonly used in the literature; the down-sampling protocol (DSP), which ignores processing delays; and the proposed embedded oriented evalution protocol (EOP), which accounts for processing delays. Example of a VOT tracker running at 7.5–10~Hz, effectively processing every fourth frame}
    \label{fig:eop}
   \end{subfigure}
   \vspace{5pt}
    
   \begin{subfigure}[c]{\textwidth}
    \centering
    \includegraphics[width=0.9\textwidth]{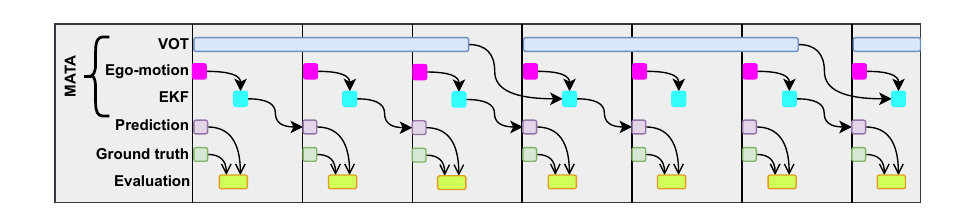}
    \caption{MATA evaluated using the EOP evaluation protocol}
    \label{fig:mata_eop}
   \end{subfigure}
   \caption{The proposed EOP evalution protocol}
    \label{fig:protocol_comparison}
\end{figure*}

The \gls{eop} protocol is presented in \cref{fig:protocol_comparison}. First, we compare  the three evaluation protocols in \cref{fig:eop} (\gls{ltp}, \gls{dsp}, and the proposed \gls{eop} protocol). In this example, the tracker processes one out of every four frames (approximately 7.5~Hz), and we aim to obtain an overall estimate of the performance it would achieve on an embedded system. In the \gls{ltp} timeline, the tracker processes every frame; as a result, the predictions closely match the ground truth. However, this scenario is unrealistic, as it does not account for computational delays. The second approach, based on \gls{dsp}, increases the motion between frames, but in practice remains similar to \gls{ltp} as the tracker still processes every frame without delay, and additional information is lost due to discarded frames. In contrast, in the \gls{eop} timeline, the processing delay is explicitly taken into account. A zero-order hold is applied to the prediction until a new output is produced by the \gls{vot} block. Consequently, this evaluation more closely reflects the actual behavior of a tracker running on an embedded system. Finally, we present the combination of \gls{mata} and \gls{eop} in \cref{fig:mata_eop}. We can compare classical \gls{vot} algorithms (\gls{eop} timeline) and the proposed tracking framework. Thanks to the asynchrony of its constituent blocks, \gls{mata} is able to generate a new prediction at every frame, thereby mitigating the computational delay inherent to the \gls{vot} model.

\subsection{Datasets}
\label{subsec:datasets}

To ensure a representative evaluation under aerial tracking conditions, we focus on datasets specifically designed for \gls{uav}-based visual object tracking. In particular, we use the UAV123 dataset~\cite{uav123}, a widely adopted benchmark for aerial tracking performance, and the VTUAV dataset~\cite{vtuav}, considering only the visible part of its test split. Although these datasets provide valuable evaluation material, they contain few sequences with occlusion~\footnote{Respectively $\thickapprox12\%$ and $\thickapprox20\%$ of number of sequences for VTUAV and UAV123}, even though \gls{uav} tracking often encounters partial or complete occlusions from vegetation, buildings, or environmental obstacles. These scenarios are crucial for evaluating trajectory estimation filters. 

\begin{figure*}[!t]
    \centering
    \begin{subfigure}[c]{0.32\textwidth}
    \includegraphics[width=0.9\textwidth]{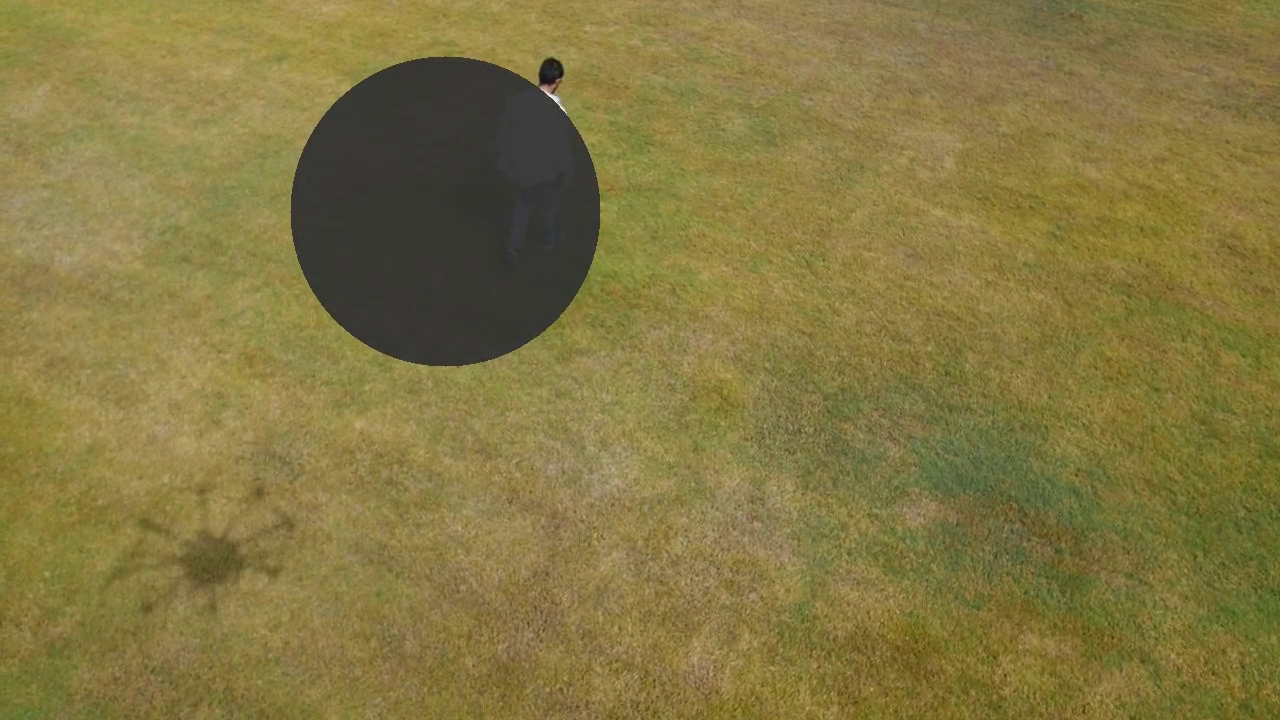}
    \caption{Circle shape}
    \label{fig:circle}
    \end{subfigure}%
    \begin{subfigure}[c]{0.32\textwidth}
    \includegraphics[width=0.9\textwidth]{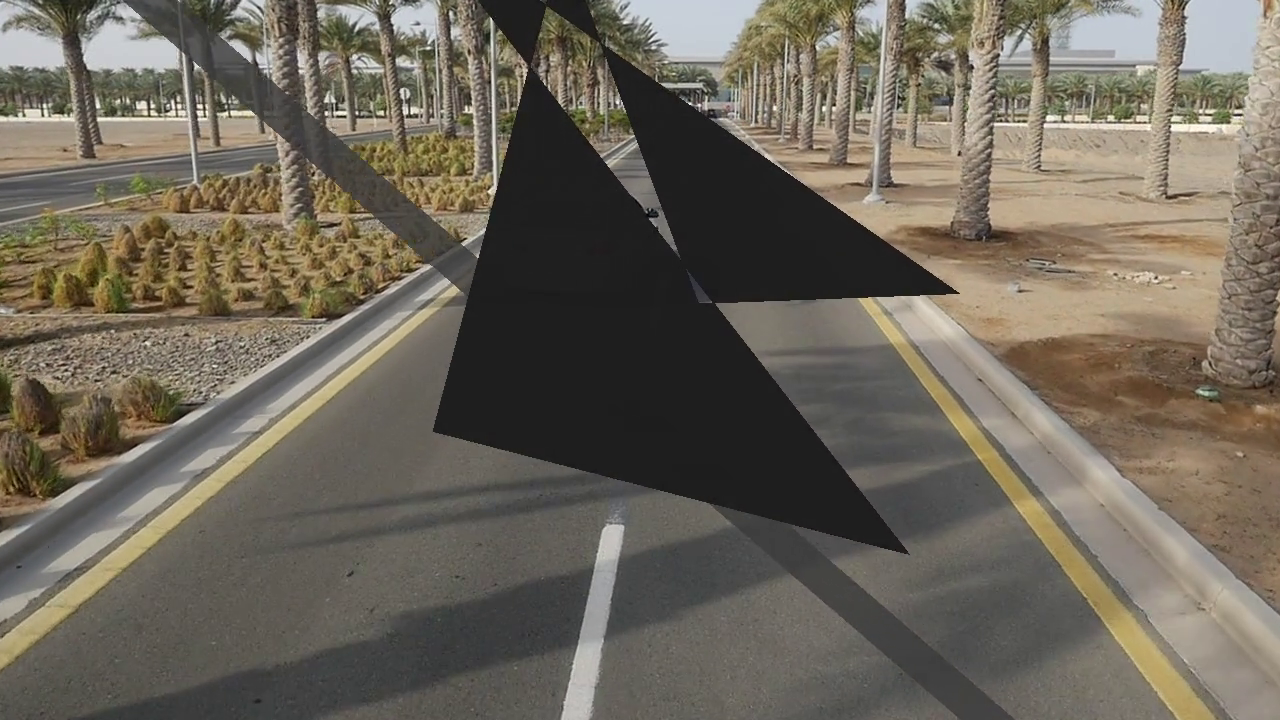}
    \caption{Polygon and stripe shape}
    \label{fig:polygon}
    \end{subfigure}
    \begin{subfigure}[c]{0.32\textwidth}
    \includegraphics[width=0.9\textwidth]{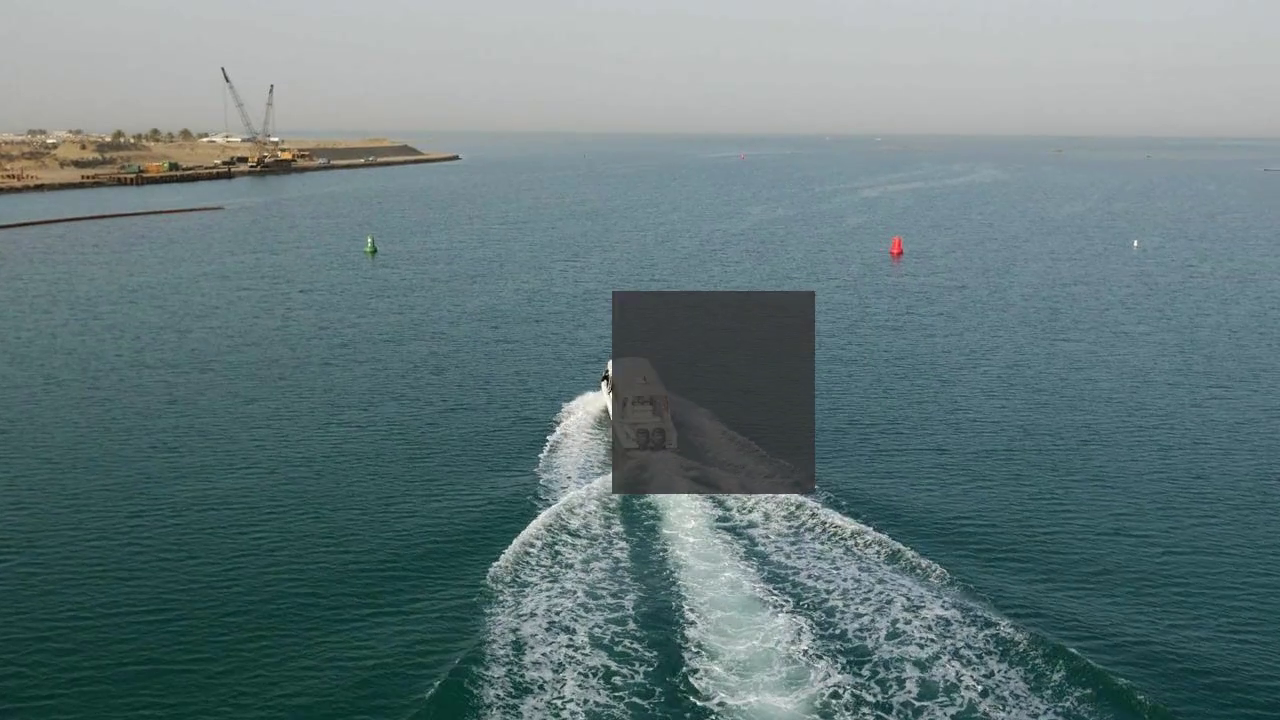}
    \caption{Rectangle shape}
    \label{fig:square}
    \end{subfigure}
    \caption{Example of shape applied to sequences on UAV123 dataset}
    \label{fig:example_aug}
\end{figure*}

To address this limitation and better evaluate the temporal reliability of \gls{mata}, we developed a data augmentation tool to generate artificial occlusions. For each frame, a binary occlusion mask is generated based on hand-crafted features such as shape, motion, and timing, specifying the start and end frames of the occlusion event. From these, the system samples start and end positions outside the image and generates a small set of “hit points” where the occlusion should intersect the object ground truth. A per-frame trajectory is then obtained by linear interpolation between these points, with small random acceleration jitter added to mimic natural motion. For each occlusion event, a single occlusion object is first selected from several geometric shapes, including rectangles (\cf \cref{fig:square}), ellipses, circles (\cf \cref{fig:circle}), blobs, or irregular polygons (\cf \cref{fig:polygon}) and its shape parameters are randomly sampled and scaled to the object size. This same object is then used throughout the entire sequence, moving along the per-frame trajectory to produce the occlusion. At each frame of the occlusion event, the binary mask of the shape is alpha-blended onto the visible frame, producing smooth occlusions. Importantly, only the overall shape and the start/end frames of each occlusion event are hand-crafted. All other parameters,  including alpha, size, exact shape parameters, number of intersection points, speed, and acceleration,  are randomly sampled, ensuring that the augmented sequences are non-deterministic, which prevents any bias in the evaluation. For fair comparison between algorithms, the random seed was fixed so that each method is evaluated on the same sequence of augmented occlusion scenarios. Ground-truth annotations are left unchanged, forcing the tracker and filter to rely on motion prediction during these intervals.

\subsection{Evaluation metrics}

To evaluate the performance of each algorithm, we adopt standard evaluation metrics commonly used in the \gls{vot} community. We are using precision rate, the distance between predicted and ground-truth centers, and \gls{sr}, the fraction of frames with sufficient bounding-box overlap. The \gls{msr} captures success across various overlap thresholds. We are using a normalized \gls{pr} to account for varying object sizes and resolutions. Computational efficiency is reported using the processing frequency (Hz), defined as the number of iterations a given block can process per second. The definitions and computation protocols for these metrics follow those presented in~\cite{borne2025military}. However, these classical metrics fail to capture how long a tracker can maintain a reliable tracking before a failure occurs. This limitation becomes particularly evident in scenarios where the tracker has no influence on the camera motion, for instance, when the camera continues to follow the object even after the tracker has lost it. In such cases, a tracker may “recover” the target purely by chance if its search region wanders erratically within the image, which artificially inflates its performance. 

To better quantify temporal persistence, several works have proposed time-linked robustness measures, such as the Mean Time Between Failures (MTBF)~\cite{7477617}, the tracking length~\cite{6ce0cefeaf7a4d0eb2455496afdc1f29}, or the robustness metric used in the VOT challenges. These metrics estimate how often a tracker fails or how long it remains on the object across entire sequences. Inspired by these approaches but aiming for a simpler and more interpretable measure, we introduce the Normalized Time-to-Failure (\gls{nt2f}) which quantify how long a tracker maintain a valid target estimate before the first failure, normalized by the sequence length. For a frame $n$ in a sequence of $N$ frames, let $R_n$ denote the predicted bounding-box region and $R_n^{gt}$ the corresponding ground truth region. The intersection-over-union is defined by $\mathrm{IoU}_n =  \frac{\lvert R_n \cap R_t^{gt} \rvert}{\lvert R_n \cup R_t^{gt} \rvert}$ with $\lvert R \rvert$ being the number of pixel in the the region $R$. Let $\tau$ be a failure threshold (in this study we use $\tau = 0$). For each sequence, the frame index of the first failure $T_f$ is computed and the sequence $NT2F$ score is obtained with \cref{eq:nt2f}.

\begin{equation}
\label{eq:nt2f}
\begin{aligned}
\mathrm{NT2F} &= \frac{T_f}{N} \\
T_f &=
\begin{cases}
\min\{ t \in [0,N-1] \mid \mathrm{IoU}_t \le \tau \},
& \text{if exists}, \\
N, & \text{otherwise}
\end{cases}
\end{aligned}
\end{equation}

\noindent At the dataset level, the $NT2F$ metric is computed as the mean of the individual $NT2F$ sequence values. This metric measures the portion of a sequence during which the tracker keeps the target before losing it for the first time. $NT2F$ is a normalized variant of the tracking length metric, focusing on how long a tracker can reliably track without external help.

\subsection{Experimental setup}

The system is evaluated on high-performance server and embedded edge platforms (respectively an Nvidia A100 GPU and an Nvidia AGX Orin developer kit) with both \gls{ltp} and the proposed evaluation protocol (\cf \gls{eop} presented in \cref{subsec:new_eval}). In this study, we employ five modular blocks: simulated camera (30~Hz), Tracking block (Block A presented in \cref{sub_sec:score}), camera ego motion compensation (Block B presented in \cref{subsec:cam_motion}) processing at 30~Hz, trajectory estimation (Block C presented in \cref{subsec:target_traj_est}) processing at 30~Hz and saving modules (30~Hz). With the exception of the tracker block, all modules operate in real time on \gls{swap} hardware, so the one-period delay is applied exclusively to the tracker. We vary the tracking frequency to characterize the system under different possible embedded configurations. We simulate tracking at speed from 5 to 30~Hz with the filter fixed at 30~Hz. After experimental trials, we observed the best results when using an ego-motion block based on a $16 \times 16$ matrix of points, uniformly distributed over the image. 

The method is validated on an NVIDIA Jetson AGX Orin representative of current edge-AI platforms. For the embedded evaluation, the system is implemented within a \gls{ros2} framework, with each block running in a separate node inside its own Docker container. Each container is allocated at least one CPU core. Blocks A and C are allocated two cores because they use a multithreaded executor to parallelize subscriber callbacks and computations. This allows block A to update the tracker state concurrently with ongoing computations, and enables block C to receive measurements from blocks A and B while running the \gls{ekf} prediction step. Communication uses the default \gls{ros2} DDS implementation. The experiment assesses system behavior under 50 W power mode. 

We evaluate two transformer-based state-of-the-art trackers, OSTrack~\cite{ye_joint_2022} and MixFormerV2~\cite{10.5555/3666122.3668683}. For the evaluation, UAV123 provides per-frame annotations, whereas VTUAV provides annotations only every 10th frame. Therefore, to compute the metrics, we subsample the predictions for the VTUAV evaluation, while using all predicted states for UAV123.

\begin{table*}[!h]
  \centering
  \begin{tabular}{c|c|c|cc|cc|cc}
    \toprule

    % Header row: dataset names with SR/NT2F label in one cell
    \multirow{2}{*}{Protocol} & \multirow{2}{*}{VOT} & \multirow{2}{*}{MATA} & \multicolumn{2}{c|}{UAV123} & \multicolumn{2}{c|}{UAV123+occ} & \multicolumn{2}{c}{VTUAV} \\
    & & & SR & NT2F & SR & NT2F & SR & NT2F \\
    \midrule
    \midrule
    % Data rows
    \multirow{4}{*}{LTP} & OSTrack & \xmark & 82.8 & 79.5 & 75.9 & 72.7 & 74.2 & 59.3 \\
    & OSTrack & \cmark & 83.6 & 79.5 & 78.5 & 74.1 & 74.8 & 61.1 \\
    & MixFormerV2 & \xmark & 84.3 & 79.1 & 78.2 & 68.8 & 75.6 & 55.2 \\
    & MixFormerV2 & \cmark & 83.2 & 79.7 & 77.2 & 72.0 & 73.9 & 59.0 \\
    \midrule
    \midrule
    \multirow{4}{*}{EOP @ 30~Hz} & OSTrack & \xmark & 80.1 & 76.9 & 73.3 & 70.1 & 74.9 & 59.4 \\
    & OSTrack & \cmark & 81.9 & 77.9 & 76.0 & 72.2 & 75.7 & 61.6 \\
    & MixFormerV2 & \xmark & 81.9 & 77.5 & 75.8 & 67.2 & 76.2 & 55.2\\
    & MixFormerV2 & \cmark & 79.1 & 78.6 & 74.5 & 73.0 & 74.9 & 60.4\\
    \midrule
    \midrule
    \multirow{4}{*}{EOP @ 10~Hz} & OSTrack & \xmark & 63.9 & 56.2 & 57.9 & 50.9 & 72.8 & 55.8 \\
    & OSTrack & \cmark & 69.8 & 64.4 & 65.8 & 60.0 & 75.5 &  58.2 \\
    & MixFormerV2 & \xmark & 65.5 & 52.2 & 60.8 & 44.5 & 74.4 & 48.4 \\
    & MixFormerV2 & \cmark & 68.5 & 64.2 & 64.7 & 60.3 & 72.6 & 56.4 \\
    \bottomrule
  \end{tabular}
  \caption{Performance of OSTrack and MixFormerV2, with and without MATA, on UAV123, UAV123+occ, and VTUAV under LTP and EOP protocols. EOP is evaluated at 30~Hz and 10~Hz, the latter matching the runtime observed on NVIDIA Jetson Orin.}
  \label{tab:result_protocol}
\end{table*} 

\section{Results}
\label{sec:results}

For \gls{eop}, we set the processing frequency of the tracking block (block A) to 10~Hz, as this corresponds to the observed processing speed on the Nvidia Jetson AGX Orin platform with the computing power set to 30~W. The same principle applies to \cref{tab:result_protocol}, \cref{tab:metrics_per_challenges}, and \cref{tab:ablation_study}.

The \cref{tab:result_protocol} compares baseline trackers and their \gls{mata} variants on three datasets using long-term evaluation protocol and the proposed \gls{eop} protocol. Our architecture consistently improves \gls{nt2f} for both trackers across all datasets. For OSTrack, it also increases \gls{sr}. For MixFormerV2 it slightly reduces SR except for \gls{eop} at 10Hz, but yields notable \gls{nt2f} gains. Overall, \gls{mata} primarily enhances temporal reliability while maintaining competitive SR performance. We observe an overall metric reduction between \gls{eop} @ 30~Hz and \gls{eop} @ 10~Hz, this shows the impact of tracker speed on performance. 

\begin{table*}[!h]
  \centering
  \begin{tabular}{c|c|cc|cc|cc|cc}
    \toprule
     \multirow{2}{*}{VOT} & \multirow{2}{*}{MATA} & \multicolumn{2}{c|}{Occlusion} &  \multicolumn{2}{c|}{Distractor} & \multicolumn{2}{c|}{High dynamic} & \multicolumn{2}{c}{Out of view} \\
     & & SR & NT2F & SR & NT2F & SR & NT2F & SR & NT2F \\
    \midrule
    \midrule
    OSTrack & \xmark & 50.2 & 43.4 & 67.8 & 60.6 & 32.0 & 16.1 & 41.2 & 39.8 \\
    OSTrack & \cmark & 59.9 & 49.9 & 74.6 & 65.0 & 38.3 & 24.0 & 47.1 & 35.9 \\
    \midrule
    \midrule
    MixFormerV2 & \xmark & 52.9 & 37.0 & 70.3 & 50.6 & 36.2 & 14.9 & 46.5 & 32.8 \\
    MixFormerV2 & \cmark & 59.1 & 50.8 & 72.4 & 61.1 & 35.1 & 25.1 & 46.4 & 36.7 \\
    \bottomrule
  \end{tabular}
  \caption{Metrics per challenges on UAV123+occ with EOP evaluation with tracker at 10Hz}
  \label{tab:metrics_per_challenges}
\end{table*} 

\noindent The \cref{tab:metrics_per_challenges} reports results on UAV123+occ at 10~Hz across different challenges. The proposed solution improves both \gls{sr} and \gls{nt2f} for OSTrack in all categories, with notable gains under occlusion and high dynamics. For MixFormerV2, our framework also improves most challenges, with a slight drop only in the out-of-view case, while maintaining comparable or higher \gls{nt2f}. Overall, \gls{mata} enhances temporal reliability under challenging conditions. 

\begin{table*}[!ht]
  \centering
  \begin{tabular}{c|ccc|c|c|c|c|c}
    \toprule
    Configuration & Ego motion & EKF & Score & Details & PR & SR & MSR & NT2F \\
    \midrule  
    \midrule
    1 & \xmark & \xmark& \xmark & MixFormerV2 (Baseline) & 79.3 & 60.8 & 51.8 & 44.5 \\  
    2 & \cmark & \cmark& \cmark & MATA & 78.1 & 64.7 & 53.5 & 60.3 \\
    \midrule
    3 & \xmark & \xmark & \cmark & section~\ref{sub_sec:score} & 78.8 & 60.8 & 51.5 & 46.4 \\ 
    \midrule
    4 & \xmark & \cmark& \cmark & no camera compensation & 76.4 & 61.6 & 51.5 & 48.1 \\    
    5 & \cmark & \cmark& \cmark & constant acceleration model & 77.9 & 64.5 & 53.3 & 57.0 \\
    6 & \cmark & \cmark& \cmark & constant jerk model & 75.4 & 62.4 & 51.6 & 54.9 \\
    7 & \cmark & \cmark& \cmark & no threshold on measurements & 77.5 & 63.1 & 52.5 & 57.6 \\
    \bottomrule
  \end{tabular}
  \caption{Impact of individual components on tracking performance on UAV123+occ, with tracking at 10~Hz using EOP.}
  \label{tab:ablation_study}
\end{table*} 

The objective of \cref{tab:ablation_study} is to demonstrate the contribution of each component of the proposed architecture: the tracking block (block A), the ego-motion compensation module (block B), and the trajectory estimation component (block C). This ablation study was conducted on the UAV123+occ dataset (\cf \cref{subsec:datasets}) using the proposed \gls{eop} evaluation protocol. As shown in this study, \gls{mata} (Configuration~2) consistently outperforms baseline MixFormerV2 (Configuration~1), particularly in \gls{nt2f}, demonstrating improved temporal reliability. Configuration~3 shows that the proposed explainable score mainly benefits long-term stability, improving \gls{nt2f} while keeping \gls{pr} and \gls{sr} comparable to the baseline. Removing camera motion compensation (Configuration~4) leads to a clear performance drop, highlighting its importance for accurate motion prediction. Alternative dynamic models using acceleration and jerk terms (Configurations~5 and~6) partially recover this loss, but remain inferior to the final constant-velocity model. Disabling measurement thresholding (Configuration~7) slightly degrades performance, confirming the benefit of filtering unreliable measurements. Overall, these results confirm that each component of the proposed \gls{mata} framework contributes to the final performance, with the estimation filter design and camera motion compensation being key factors for temporal reliability. 

\begin{table*}[!h]
\centering
\begin{tabular}{c|c|ccc||ccc|ccc}
\toprule
\multirow{3}{*}{VOT} & \multirow{3}{*}{MATA} 
& \multicolumn{3}{c||}{Baseline} 
& \multicolumn{6}{c}{Absolute difference w.r.t. Baseline} \\

& & \multicolumn{3}{c||}{Embedded (\gls{ros2})}
& \multicolumn{3}{c|}{EOP} 
& \multicolumn{3}{c}{LTP} \\

& & SR & NT2F & Freq. 
& $\Delta$SR & $\Delta$NT2F & Freq. 
& $\Delta$SR & $\Delta$NT2F & Freq. \\

\midrule
\midrule

OSTrack & \xmark & 58.6 & 51.0 & 10.5 & 3.7 & 4 & 10.5 & 17.3 & 21.7 & 30 \\
OSTrack & \cmark & 58.4 & 51.3 & 9.2 & 0.5 & 1.6 & 9.2 & 20.1 & 22.8 & 30 \\

\midrule
\midrule

MixFormerV2 & \xmark & 63.8 & 51.0 & 11.1 & 5.7 & 8.5 & 11.1 & 14.4 & 17.8 & 30 \\
MixFormerV2 & \cmark & 59.7 & 49.8 & 9.7 & 1.0 & 3.8 & 9.7 & 17.5 & 22.2 & 30 \\

\bottomrule
\end{tabular}

\caption{Embedded performance on UAV123+occ (Jetson AGX Orin). EOP and LTP report absolute differences ($\Delta$) in SR and NT2F with respect to the Embedded baseline. For example, the absolute difference in SR between EOP and the baseline is computed as $\Delta SR = \left| SR_{\mathrm{EOP}} - SR_{\mathrm{baseline}} \right|$. Freq. denotes the processing frequency in Hz.}
\label{tab:embedded_results}
\end{table*} 

\noindent The \cref{tab:embedded_results} reports the performance of tested trackers on the NVIDIA Jetson Orin AGX using the UAV123+occ dataset. The \gls{eop} and \gls{ltp} columns report the absolute differences of \gls{sr} and \gls{nt2f} between server evaluations and embedded system results. \gls{eop} accounts for the tracker’s actual embedded speed, whereas \gls{ltp} does not. We observe that the \gls{eop} protocol better reproduces embedded performance than \gls{ltp}, reducing the average discrepancy with the embedded baseline by about $6.4\times$ for \gls{sr} and $4.7\times$ for \gls{nt2f} across the tested methods, indicating that \gls{eop} better captures the effects of real-time and asynchronous processing on resource-constrained hardware. Interestingly, the \gls{mata}-enhanced architecture no longer provides clear improvements in the embedded setup, suggesting that its benefits may be diminished under actual hardware constraints (hypotheses for this are discussed in \cref{sec:discussion}).

\section{Discussion}
\label{sec:discussion}

The results show that \gls{mata} enhances temporal reliability across diverse conditions, while the \gls{eop} protocol provides a more realistic assessment of embedded tracking performance under real-time constraints.

The experimental results confirm that \gls{eop} offers a closer approximation of embedded tracking performance than \gls{ltp}. When tracker speed is correctly configured, \gls{eop} reproduces embedded trends well, making it a valuable tool for early-stage embedded system design. In contrast, \gls{ltp} ignores real-time constraints and therefore overestimates achievable performance. Compared to UAV123, the relatively low variation in VTUAV performance across \gls{ltp}, \gls{eop} @ 30 Hz, and \gls{eop} @ 10 Hz is likely due to sparse annotations. This sparsity leads to the loss of fine-grained ground-truth object states, especially important with \gls{sr} where traditional trackers might recover the object and inflate the \gls{sr} value. However, the performance degradation is more pronounced in term of \gls{nt2f}. We also observe a remaining gap remains between \gls{eop} and embedded results. This discrepancy mainly arises from \gls{eop}’s assumption that computation latency dominates, whereas in \gls{ros2} systems, communication latency plays a significant role. Even real-time-capable modules, such as ego-motion estimation, may be slowed by inter-node communication. In practice, each block is subject to three types of latency: processing delay (blocking), waiting delay (e.g., awaiting new input), and communication delay (non-blocking but cumulative). These latencies are highly variable and non-deterministic on embedded platforms, unlike the constant delays assumed in \gls{eop}.

The proposed tracking architecture improves performance over the baseline with some limitations. Under \gls{ltp}, it increases \gls{nt2f} while maintaining comparable SR, demonstrating improved long-term stability. Under \gls{eop}, its advantage grows as tracker speed decreases, confirming its effectiveness for low-frequency tracking and occlusion scenarios. However, \gls{mata} relies heavily on accurate ego-motion compensation, as confirmed by the ablation study. The remaining performance gap between \gls{eop} and embedded evaluations highlights limitations of the current implementation. Since our framework was tuned under \gls{eop} assumptions, delayed ego-motion estimates in the embedded system directly reduce its effectiveness, indicating a primarily engineering-related limitation. Moreover, the current \gls{mata} formulation assumes instantaneous measurements, which does not hold in current \gls{ros2} implementation. Future work will therefore focus on reducing communication latency and extending the framework to explicitly model and compensate for measurement delays.

The current ego-motion block (block B) of \gls{mata} has some limitations. It is currently based solely on visual information, which is suboptimal because vision can fail in many scenarios. For instance, performance can degrade at night, in low-light or foggy conditions, or when the camera is subject to motion blur. Its accuracy is also limited by the frame rate and depends on sensor characteristics such as pixel size, field of view, and lens distortion. Other ego-motion estimation methods are not perfect either, as discussed in \cref{sub_sec:score}. One potential solution is to integrate visual-inertial odometry (VIO), which fuses visual and inertial measurements to improve robustness and accuracy. Future work will explore this direction to enhance the performance of the \gls{mata} method.

\section{Conclusion}
\label{sec:conclusion}

In this study, we address two key challenges for embedded \gls{uav} applications of \gls{vot} solutions. First, we tackle the lack of temporality in current trackers and the trade-off between tracking accuracy and embedded system constraints. Second, we focus on the evaluation of trackers in a manner that reflects realistic embedded deployment conditions. To this end, we consider both architectural improvements for robust tracking and the development of evaluation tools better suited to the limitations and characteristics of embedded \gls{uav} platforms.

To tackle these issues, we propose a new tracking architecture which is the \gls{mata} framework, integrating ego-motion compensation and an estimation filter with modern deep trackers. This design enables the tracker to explicitly account for both camera motion and object dynamics, improving robustness in challenging aerial scenarios while balancing performance and inference speed. To evaluate this architecture, we developed a data augmentation tool to simulate object occlusions, resulting in an augmented version of the UAV123 dataset, named UAV123+occ. Experiments conducted on this dataset demonstrate that the proposed framework improves temporal reliability. We also design a new metric suitable for scenarios where tracker re-initialization may not be possible (\gls{nt2f}). Finally, we introduce a new evaluation protocol which accounts for processing delays, namely the embedded-oriented evaluation protocol (\gls{eop}).  It further highlights the benefits of the proposed framework under realistic processing constraints, particularly with respect to the \gls{nt2f} metric. 

The current \gls{mata} implementation is limited by its reliance on visual-only ego-motion, sensitivity to communication and processing latencies, and assumptions of instantaneous measurements, pointing to future work on visual-inertial integration and explicit modeling of system and sensor delays.

\medskip

% Acknowledgements
\medskip
\section*{Acknowledgments}
This work is financed by the Agence de l'Innovation de Défense (AID) from the french ministry of defense.

\section*{Conflict of Interest}

The authors declare that they have no financial or commercial conflicts of interest.

\section*{Data availability statement}

The datasets used in this study are publicly available benchmark datasets. Details and references for each dataset are provided in the manuscript. Any additional data generated during the experiments can be made available by the authors upon reasonable request.

\printglossary[type=\acronymtype]

\bibliographystyle{ieeetr}
\bibliography{
     biblio
}

\end{document}